# Generalizable and Robust Deep Learning Algorithm for Atrial Fibrillation Diagnosis Across Ethnicities, Ages and Sexes

Shany Biton[1], Mohsin Aldhafeeri[2], Erez Marcusohn[3], Kenta Tsutsui[4], Tom Szwagier[5], Adi Elias[3], Julien Oster[6], Jean Marc Sellal[2], Mahmoud Suleiman[3], and Joachim A. Behar[1*]

[1]Faculty of Biomedical Engineering, Technion-IIT, Israel
[2]Department of Cardiology, Centre hospitalier Universitaire de Nancy, France
[3]Department of Cardiology, Rambam Medical Center and Technion The Ruth and Bruce Rappaport Faculty of Medicine, Haifa, Israel
[4]Department of Cardiovascular Medicine, Faculty of Medicine, Saitama Medical University International Medical Center, Japan
[5]Mines Paris, PSL Research University, Paris, France
[6]IADI UMR 1254, INSERM, Universite de Lorraine, Nancy, France

## Abstract

To drive health innovation that meets the needs of all and democratize healthcare, there is a need to assess the generalization performance of deep learning (DL) algorithms across various distribution shifts to ensure that these algorithms are robust. This retrospective study is, to the best of our knowledge, the first to develop and assess the generalization performance of a deep learning (DL) model for AF events detection from long term beat-to-beat intervals across ethnicities, ages and sexes. The new recurrent DL model, denoted ArNet2, was developed on a large retrospective dataset of 2,147 patients totaling 51,386 hours of continuous electrocardiogram (ECG). The model's generalization was evaluated on manually annotated test sets from four centers (USA, Israel, Japan and China) totaling 402 patients. The model was further validated on a retrospective dataset of 1,730 consecutives Holter recordings from the Rambam Hospital Holter clinic, Haifa, Israel. The model outperformed benchmark state-of-the-art models and generalized well across ethnicities, ages and sexes. Performance was higher for female than male and young adults (less than 60 years old) and showed some differences across ethnicities. The main finding explaining these variations was an impairment in performance in groups with a higher prevalence of atrial flutter (AFL). Our findings on the relative performance of ArNet2 across groups may have clinical implications on the choice of the preferred AF examination method to use relative to the group of interest.

**Funding:** Grant (3-17550) from the Ministry of Science & Technology, Israel & Ministry of Europe and Foreign Affairs (MEAE) and the Ministry of Higher Education, Research and Innovation (MESRI) of France and grant ERANET - 3-16881 from the Israeli Ministry of Health.

## 1. Introduction

Atrial fibrillation (AF) is the most prevalent heart arrhythmia[1,2]. It is associated with a 5-fold increase in stroke incidence and a 3.5-fold mortality risk increase[3]. Across ethnicities, ages and sexes AF has been reported to have varying prevalence[1,2,4–6]. A recent Swedish study[7] reported on 89 out of 908 (9%) false positive automated misdiagnoses of AF, with almost half of these cases not corrected by the overreading primary-care physician, sometimes leading to inappropriate treatment with anticoagulant therapy. Machine learning (ML) presents a unique opportunity to provide an accurate automated diagnosis of AF. Yet, these models must demonstrate generalizability to external datasets integrating a range of population samples. We searched PubMed up to May 24th, 2022, for research articles containing the terms "(deep learning or convolutional neural network)" AND "(atrial fibrillation or atrial flutter)" AND "(generalization across ethnicity or generalization across age or generalization across sex)", without any date or language restrictions. We also reviewed reference lists of eligible articles. Our search did not identify any previous studies assessing deep learning (DL) algorithms generalization across ethnicities, ages and sexes for the task of atrial fibrillation (AF) detection. This work addresses the challenge of generalization performance of a novel DL AF event detection algorithm for these groups.

## Results

### Datasets

The University of Virginia dataset (UVAF) from the USA[8,9] was used to train the ML models. UVAF consists of a dataset of 2,147 patients totaling 51,386 hours of continuous electrocardiogram (ECG). The model's generalization was then evaluated on manually annotated events in test sets from four centers (USA, Israel, Japan and China) totaling 402 patients. These test datasets are denoted UVAF-test, RBDB-test, SHDB-test, CPSC-test, (Table 1). Finally, the new DL model was validated on a retrospective dataset of 1,730 consecutives Holter recordings from the Rambam Hospital Holter clinic, Haifa, Israel (denoted RBDB-test2). Overall, the datasets totaled 4,298 ECG Holter recordings and over 99,705 hours of continuous data. Annotations for AF and Atrial Flutter (AFL) were grouped under a single label denoted $AF_l$ for the experiments and similar to previous works[8,9,12,13]. Holter recordings of patients under the age of 18 years and corrupted recordings were excluded. Overall, out of 2,891 UVAF-train recordings, 106 (3.7%) were excluded due to low quality or missing annotations as summarized in Figure 1. No recordings were excluded from the test sets. The PhysioZoo software[15,16] was installed on a server within the Technion domain and accessed remotely by a resident cardiologist annotator (MA), who annotated the test set Holter ECGs. Supraventricular arrhythmias were divided and annotated, using our annotation protocol, into five categories: (1) AF, (2) AFL, (3) atrial tachycardia (AT), (4) Other supraventricular tachycardias such as Wolf-Parkinson-White and intranodal tachycardias and (5) other, such as NSR, that were not labelled. Reading time by the resident (MA) was estimated to be on average 45 min per 24 h Holter recording.

### Deep learning

ArNet2 builds on our previous work developing ArNet[12]. ArNet2 works in two steps (Figure 2): the input to the first step is a 60-beat RR-interval window ($w_s$), i.e., 59 RR intervals, and the output is a binary classification of $AF_l$/non-$AF_l$. The Deep Learning (DL) architecture is divided into two parts (Figure 2): a first part constituted of residual blocks and used to build representation of an ECG signal window, thereby creating an imbedding. The second step consists of a sequence encoder made of GRU units to take the time representation into account. ArNet2 was benchmarked against ArNet and a gradient boosting (XGB) model taking as input heart rate variability (HRV) features[12]. The following statistics were computed to assess model performance for the individual 60-beat-window classification task: sensitivity (Se), specificity (Sp), positive predictive value (PPV), AUROC and the harmonic mean between Se and PPV (F1). In addition, the absolute difference between the estimated and reference AF Burden (AFB) denoted $|E_{AF}(\%)|$ was computed as an overall clinical performance measure.

**Overall model performance**
The best UVAF-test set score for window classification was obtained for ArNet2 (Figure 3 and Table S2) with an F1 of 0.95 compared to F1 of 0.93 for ArNet and 0.92 for XGB. The $|E_{AF}(\%)|$, for the UVAF-test set was 0.1 (0.0-2.41) for ArNet2, 0.08 (0.0-3.29) for ArNet and 1.42 (0.3-5.31) for XGB. Figure S2 shows the histogram of $|E_{AF}(\%)|$ for each $AF_l$ severity level for ArNet2. A proportional T-test revealed that the F1 performance was significant between ArNet2 and ArNet (p<0.001). A paired T-test revealed that the gap in $|E_{AF}(\%)|$ was significant (p<0.001) between ArNet2 and ArNet.

**Performance across ethnicities**
Table 2 and Figure 3A present performance statistics for ArNet2 across different ethnic groups. When considering F1, ArNet2 consistently outperformed other ML models (Table S2 and Figure 3A) with all external test sets apart for the CPSC-test set, where ArNet2 scored an F1 of 0.95 compared to F1 of 0.95 for ArNet and 0.94 for XGB. The range of ArNet2 performance across the different ethnic groups varied in the range of F1=0.90-0.95, with the best performance obtained for the CPSC-test set and the lowest for the RBDB-test set. The median performance in terms of $|E_{AF}(\%)|$ was similar between ArNet2 and ArNet, but the Q3 was systematically smaller for ArNet2.

**Performance across sexes**
Table 2 and Figure 3B present results for ArNet2 across different sex groups. ArNet2 performed significantly better (p<0.001) on recordings of female (F1=0.94) compared to male (F1=0.90) patients. The ArNet2 $|E_{AF}(\%)|$ for the female group was 0.14 (0.0-2.46) compared to 0.71 (0.06-4.79) for the male group. This relative trend was also observed for the other models (XGB and ArNet) as well (Table S3 and Figure 3B).

**Performance across age groups**
Table 2 and Figure 3C present performance measures for ArNet2 across different age groups. ArNet2 performed better on recordings of adult patients aged ≤60 years (F1 of 0.95) compared to patients >60 years (F1 of 0.90 for ages 61-75 years and 0.94 for >75 years). Similarly, the $|E_{AF}(\%)|$ for the ≤60 years age group was smaller (0.14 (0.0-2.81)) than for the older age groups (0.47 (0.05-5.15)) for 60-75 years and (0.22 (0.0-2.66)) for >75 years. Similar performance was measured for the other models (XGB and ArNet) as well (Table S4 and Figure 3C).

**Performance for simulating intended use scenario**
In the simulated intended use scenario and while considering a fully automated AF diagnosis algorithm based on ArNet2 and setting a decision threshold on the AFB at 2% (Figure 4), the algorithm yielded a PPV of 0.57, NPV of 0.99, Se of 0.92 and Sp of 0.95 at the patient level. ArNet2 performed better for the female group, with a PPV of 0.73, NPV of 0.99, Se of 0.94 and Sp of 0.97 compared to the male group which obtained a PPV of 0.45, NPV of 0.99, Se of 0.9 and Sp of 0.93.

**Combined test set error analysis**
Out of 300 combined manually annotated test set recordings, 209 patients had AF events and no AFL events, 5 patients had AFL events and no AF events and 14 had a mixture of AF and AFL events. In total, 69.53% of the AFL windows were misclassified as non-$AF_l$ by ArNet2. This represented 47.66% of all False Negatives (FN) and reflects the fact that ArNet2 performs poorly in detecting AFL events. This led to underestimation of the $|E_{AF}(\%)|$ which may result in the misclassification of these patients as non-$AF_l$. In parallel, 85.72% of the FP windows were labelled as other rhythms,

8.76% as AT, 4.91% as mixed labels and 0.6% as atrial bigeminy (AB) (see Figure S3). Examples of misclassified windows are shown in Figure S4.

**Simulating intended use scenario**

Using ArNet2 on the RBDB-test2 with a threshold at 2% on the AFB we obtain PPV of 0.57, NPV of 0.99, Se of 0.92 and Sp of 0.95 in performing diagnosis. Of the FP patients, 37.93% were diagnosed with other coronary artery diseases (CAD), including myocardial infarction and congestive heart failure. For the FN cases, the cumulated duration of AF events was very small with 56.82, (33.54-77.26) sec. The AFB distribution is shown in Figure 4. A total of 87 FP cases were reviewed by a cardiologist; the most prevalent arrhythmias that were misclassified as $AF_l$ were AT (20.68%), sinus arrhythmia (19.54%) and premature ventricular contractions (17.24%).

## 2. Discussion

The main contribution of the research are the methods and results to assess the generalization performance of a novel DL algorithm for AF events detection across different ethnic groups, sexes and ages. Generalization performance was evaluated using four test sets from different countries (USA, Japan, Israel and China) with two of the test sets (Israel and Japan) that were newly elaborated for this research. ArNet2 generalization performance across ethnic groups was consistently higher than that of the benchmark algorithms (Figure 3A). ArNet performance across ethnic groups varied from an F1 of 0.90 for RBDB to 0.95 for CPSC. AFL events were identified in 2.94% of the UVAF, 7.22% of the SHDB, 14.13% of the RBDB and 0.43% of the CPSC recordings. When assessing ArNet2 across sex, performance was higher for females than males. The vast majority (83.34%) of AFL windows in the combined test set belonged to the male group and only 16.66% belonged to the female group. Thus, the prevalence of AFL among men was significantly higher than in women in our test set. When assessing ArNet2 for different age groups, the percentage of AFL windows out of the total $AF_l$ windows per age group was found to be 0.8% for adults aged ≤60 years, 10.5% for adults aged 60-75 years and 6.3% for the ≥ 75 age group. Taken together, the relative prevalence of AFL to AF was significantly higher in the elderly population (> 60 years). The observations made for ArNet2 performance across ethnicities, sexes and age groups were consistent for XGB and ArNet (Figure 3).

The observations on performance presented by patient sex are consistent with a previous single-center study conducted by a group from Wisconsin (USA)[22], who analyzed medical records from a dataset of 58,820 patients. The authors reported that AFL was 2.5 times more common in men ($p < 0.001$). The overall incidence of new cases of AFL during the four-year study period was 88 per 100,000 person-years, ranging between 5 per 100,000 in patients aged <50 years and 587 per 100,000 in patients >80 years of age. Since models working from beat-to-beat intervals can hardly detect AFL windows, we suggest that the higher prevalence of AFL windows found in RBDB (versus other ethnic groups), male (versus female) and aged (versus younger) groups explains the lower performance of ArNet2 for these subpopulations. These observations may have important clinical implications suggesting, for example, that for the male population, detection of $AF_l$ events may necessitate the use of a portable device measuring continuously the raw ECG waveform. However, there is limited research comparing $AF_l$ events detection based on beat-to-beat intervals versus on raw ECG data and thus it remains to be experimentally proven that analyzing the raw ECG versus the beat-to-beat interval variation would yield better results. Taken together, epidemiological consideration on the relative prevalence of AFL to AF across different ethnic groups and across age groups is warranted to better define the population sample expected to benefit from data-driven models based on the beat-to-beat interval variation versus the raw ECG waveform.

The second main contribution of this research is the elaboration of ArNet2, a new robust DL model for $AF_l$ events detection from long term beat-to-beat intervals time series. In the context of AF diagnosis based on the irregularity of the RR-interval, notable research include single features drawing from information theory such as the coefficient of sample entropy (CosEn)[20] or Lorenz-based features[21]. We have shown the superiority of ArNet2 over classical ML approaches taking engineered features such as CosEn as input[12]. ArNet2 also significantly outperformed (p<0.001) ArNet (Figure 3 and Table S2-3).

The third main contribution of this research was the evaluation of ArNet2 on a large dataset of 1,726 consecutive recordings collected from Rambam Hospital Holter clinic. The model better performed with female (PPV=0.73, NPV=0.99) than male (PPV=0.45, NPV=0.99) recordings, consistent with the window classification results obtained using the combined test set. Among the 1,601 non-AF recordings, a total of 87 would have been flagged as AF, subsequently requiring human review of the respective ECG recordings and thus incurring some cost. While the intention of the automated data driven analysis is to achieve the highest PPV possible, among these 87 recordings, 63 had another cardiac abnormality and thus review of their ECG by a qualified health professional would have been worthwhile anyway. This further highlights the perspective of our robust data-driven algorithm. Among the 125 patients with AF, a total of 115 were correctly diagnosed with AF in a fully automated manner. For the FN cases, the cumulated duration of AF events was very small with 56.82, (33.54-77.26) sec. Thus, although these individuals had AF events it is not certain that they would have been treated for it given the short episodes. In addition, ArNet2 accurately estimated the AFB, which is an important statistic when considering treatment options and dosage. For example, recent data suggest anticoagulation treatment for AFs persisting for more than 6 minutes provides no clinical benefit[23].

This research had several limitations. First, generalization was evaluated on a limited number of Holter recordings (402 overall) and for a limited number of ethnic groups (four different countries) without elaboration on racial diversity. Although the performance of ArNet2 was high and outperformed benchmark algorithms, it failed to detect a significant proportion of AFL windows (47.66%), resulting in variation in performance across ethnicities, ages and sexes. This is an intrinsic limitation of a data-driven algorithm working on beat-to-beat interval variation versus raw ECG data. When detecting $AF_l$ events, 8.7% of FPs were labeled as AT rhythms. This is an intrinsic limitation of working with retrospective data since distinguishing between AT and AFL is extremely challenging for a cardiologist (without vagal maneuver or/and adenosine test). Another 4.91% of FPs were mixed-label but with a non-AF dominant rhythm. This is also an intrinsic limitation of the methodology, since we work using 60-beat windows. Finally, 85.51% of FPs were labelled as other rhythms. More work is needed to understand the reasons for misclassification of these windows.

In the present work, and similar to previous research on AF detection using ML, AF and AFL events were grouped into a single $AF_l$ class. However, distinguishing between AF and AFL has clinical impact on the diagnosis and management of patients with one or the combination of these atrial arrhythmias. Unlike AF, in which rhythm control and rate control are reasonable strategies, maintenance of sinus rhythm is desirable in most patients with AFL to control symptoms. For most AFL patients, radiofrequency catheter ablation is preferred over pharmacologic therapy because of the high success rate and low rate of complications. In addition, many patients with AFL have episodes of AF. Simultaneous ablation of AF and cavotricuspid isthmus (CTI)-dependent AFL can be performed when both arrythmias are recorded before the procedure. However, the best approach for patients referred for ablation with CTI-dependent AFL, without history of AF has not been defined[24,25]. In conclusion, because AF and AFL are managed differently, with the exception of anticoagulation therapy, a significant proportion of patients would benefit from the diagnostic tools distinguishing between the two conditions. Although the incidence of AFL in the general population is still uncertain, the analysis of Granada et al.[22] estimated it to be 200,000 new cases per year in the United States. There is thus a strong motivation to create data-driven algorithms that can detect and distinguish between AFL and AF events. For this purpose, the development of data-driven models based on raw ECG data may be necessary.

To drive health innovation that meets the needs of all and democratize healthcare, there is a need to assess the generalization performance of deep learning (DL) algorithms across various distribution shifts to ensure that these algorithms are robust. Overall, the datasets for our analysis totaled 4,298 recordings and over 99,705 hours of continuous data. We demonstrated that our new model is robust and generalizable across ethnicity, ages and sexes being consistently significantly superior to benchmarks. We attributed differences in performance across ethnic groups, gender and age to the relative prevalence of AFL. Since AF and AFL have different clinical management pathway, our findings support the need to discriminate between the AF and AFL classes in future work. Research is also warranted in documenting the relative prevalence of AFL to AF according to age as well as across ethnicities. The algorithm developed can be used to power remote monitoring wearable solutions such as smartwatches or single lead ECG patches. This robust data-driven capability imbedded in wearables opens to the perspective of long term continuous remote monitoring for diagnosis of AF in groups such as those with embolic stroke of undetermined source or to follow-up on ablation therapy in those at risk for recurrent AF.

## 3. Methods

**Development set**

The UVAF[8,9] consists of ECG Holter recordings of patients for whom the University of Virginia health system physicians ordered Holter monitoring between December 2004 to October 2010. Indications for the Holter recordings included palpitations (40%) or syncope and dizziness (12%). The dataset contains 2,247 annotated recordings of individual patients over the age of 18 years. Recordings were digitized at 200Hz. UVAF was used to train (UVAF-train, n=2,147) and evaluate (UVAF-test, n=100) the ML models.

**Test sets**

The demographic characteristics of patients from the test sets are summarized in Table 1 and Figure S1. We elaborated three test sets denoted UVAF-test, RBDB-test and SHDB-test. These include two newly created datasets: the Rambam Hospital dataset (RBDB) from Israel (IRB: D-0402-21) and the Saitama Hospital dataset (SHDB), Japan (IRB: 20-030). For that purpose, a total of 100 recordings were selected from the original databases while stratifying by age, sex and diagnosis for $AF_l$. Specifically, 80 recordings labeled as $AF_l$ and 20 recordings labeled as non-AF were randomly chosen for each test set. For the RBDB-test and SHDB-test, the $AF_l$ diagnoses were obtained from the medical report prepared following the patient's examination. As no patient reports were available for UVAF, the label was inferred from the $AF_l$ events annotated in the recordings. For the CPSC-test the complete available open dataset was used as is. The CPSC-test[10,11] contains 1,136 variable-length ECG recordings extracted from lead I and lead II of 102 long-term dynamic ECGs, digitized at 200Hz. The variable-length ECG recordings fragments were grouped into patient recordings using the ID numbers located in the filename. The RBDB-test consists of 100 ECG recordings of adults for whom physicians from multiple hospital departments ordered Holter monitoring between October 2013 and November 2021. Holter recordings from the RBDB-test were digitized at 128Hz, with 12-bit resolution over a 10mV range. The SHDB-test consists of 100 adult ECG recordings of patients for whom treating physician ordered Holter monitoring between November 2019 and January 2022. Holters from the SHDB-test were digitized at 125Hz. The UVAF-test, RBDB-test and SHDB-test recordings were manually re-annotated (see section 2.1.4). For the CPSC-test the open-access annotations were used as is.

**Simulating intended use scenario**

In addition to the RBDB-test recordings, an additional 1,726 consecutive recordings retrospectively obtained from the clinic from the period August 2017 and November 2021, were obtained from the Rambam Hospital Holter clinic. This dataset is denoted RBDB-test2. The free texts of the Holter examination reports were reviewed manually for the diagnosis of AF and AFL. If short AF or AFL events were explicitly reported, then the corresponding Holters were reviewed by a resident cardiologist (AE) and if the event(s) were shorter than 30 s, then a non-AF label was given to the recording. If at least one event was longer than 30 s, then an $AF_l$-positive label was given to the recording[14]. As a result, the RBDB-test2 set consisted of 1,726 recordings with a binary label for the presence or absence of $AF_l$ rhythm. This subset of RBDB was used to simulate the results of the new algorithm in performing a fully automated diagnosis on an intended use population sample.

**Expert annotations of test sets**

The Holter ECG makes it possible to record the cardiac electrical activity continuously over 24 hours or longer. It helps physicians to detect arrhythmic disorders. In annotating AF events for this work the heart rate (HR) time series, RR time interval variability and QRS complexes were mainly used[26,27]. The HR time series enables to assess whether the HR is normal defined as within the interval of 60 to 100 bpm or in tachycardia defined as an HR over 100 bpm. When a tachycardia was detected, it was determined if this was sinus rhythm or not and the type of tachycardia was annotated. Yet, some AF events happen during periods of normal or low HR. This can be the case in patients receiving treatments lowering HR (betablockers). AF is a rhythm disorder presenting irregular variation in the RR time series. This irregularity enables to detect AF events even at a normal HR. Irregularities of the RR time series may also be due to the detection of atrial, junctional and ventricular extrasystoles. However, these extrasystoles are easily recognizable and do not represent a diagnostic challenge. Thus, it is possible to effectively detect AF events occurring during normal HR. AFL is defined as a disorder of regular supraventricular rhythm due to a macro reentry in right atrium, recognized with the presence of F-wave on ECG. The distinction between AF and AFL is generally easy. However, in the event of an AFL with a HR faster than 150 bpm, the F-wave are often not visible, and it can pose a diagnostic challenge. In hospital, the vagal maneuver or the adenosine test are performed in order to slow down the HR to better see the atrial activity and distinguish the F-wave. However, this is infeasible in working with retrospective recordings. The distinction between AF and AFL at a high HR thus remains a limitation. In supraventricular rhythm disorder QRS complexes are less than 120 ms. However, in some instances, patients can present wide QRS, when a bundle branch is present. It can be permanent (organic bundle branch) or occurring only for high HR (functional bundle branch block). This may mimic ventricular tachycardia, characterized by wide QRS tachycardia. However, in AF the rhythm is irregular which enables a discriminative diagnosis between AF and ventricular tachycardia in these instances. Using this protocol, supraventricular arrhythmias were annotated into four categories: (1) AF; (2) AFL; (3) atrial tachycardia (AT); (4) Other supraventricular tachycardias such as Wolf-Parkinson-White and intranodal tachycardias and (5) other such as NSR that where not labelled. No semi-automated analysis or technician was used to pre-identify AF events. Rather the resident worked from the raw Holter recording with information on instantaneous HR and RR interval variation being pre-computed by the software.

**Machine learning for AF events detection**

*ArNet2 architecture*

The first part is formed by stacking $n_b$ of 5 residual blocks that consists of two 1D convolutional layers per block, with a Batch Normalization (BN) layer and Rectified Linear Unit (ReLU) adopted from He et al.[28] prior to each convolutional layer. A dropout layer with rate ($d_{r1}$) of 0.2 is added between the blocks with a shortcut connection and Max Pooling. Every two blocks, the number of filters from one layer to the next doubles while the input length is divided by two, with initial number of filters ($n_f$) of 64 and a filter length ($f_l$) of 10. The final block is composed of Flattening and three successive Dense and Dropout layers with dropout rate ($d_{r2}$) of 0.5, at this point, the embedded features are extracted. The initial number of hidden layers ($n_{h_u}$) is 512, being divided by two each time. After the extraction, a last Dense layer of size 1 outputs a binary classification for each input window. The chosen loss for the first training step is a weighted binary cross-entropy loss, giving higher importance to the windows labelled as $AF_l$, and optimized with Adam[29] with a learning rate $\alpha$ of $10^{-2}$. Then, in the second step, a Recurrent Neural Network (RNN) takes as input the embedded features as well as the corresponding classification output from the previous step. This RNN consists of 4 gated recurrent units (GRU) with the same architecture, two dense layers and a sigmoid function which acts as an activation function and outputs a probability label of $AF_l$ for each window. Each GRU is trained with windows belonging to a target population (Non-AF, $AF_{mild}$, $AF_{mod}$ and $AF_{sev}$) to better capture the temporal distribution of AF events for a given AFB corresponding to a given severity level, as defined in Chocron et al.[12]. For each window, the embedded features from the 9 preceding windows (when available, $h$) are concatenated along the time axis. Then, one GRU unit is selected out of the four based on the patient's AFB and fed with the resulting features array. The output of the model based on the GRU unit is a classification of label per window. During training, the AFB used is the true AFB. However, during inference, the estimated AFB based on the first step output classification is used.

*Training strategy*

The UVAF training recordings (UVAF-train) were divided into 60-beat windows (**Figure 1**). For each window, a label, i.e. $AF_l$ or Non-$AF_l$ was assigned, based on the most prevalent beat label in the 60-beat window. ML models were trained for binary classification of individual windows as $AF_l$ or non-AF. Hyperparameters were optimized and selected using Bayesian search[17] using 5-fold cross-validation. The hyperparameters search space is described in Table S1. The proposed network ArNet2 was benchmarked against a gradient boosting (XGB) model and ArNet developed in our previous work[12]. The models were optimized to maximize the area under the receiver operating characteristic (AUROC) on the validation set and the threshold on the output probabilities were defined as the point which maximized the F1 on the training set. Models were trained on the UVAF-train and evaluated on the internal test set (UVAF-test) as well as the external test sets (RBDB-test, SHDB-test, CPSC-test). Models were built using TensorFlow 2.4 and trained using a single NVIDIA A100 GPU.

**Performance statistics**

The following statistics were computed to assess model performance for the individual 60-beat-window classification task: sensitivity (Se), specificity (Sp), positive predictive value (PPV), AUROC and the harmonic mean between Se and PPV (F1).

A clinical performance measure, namely the AF Burden (AFB), which we defined as the percentage of time spent in $AF_l$. Research has suggested superior prognostic value of AFB for stroke[18] compared to a binary diagnosis. Patients were categorized into four groups based on their AFB as defined in our previous work[12]: Non-AF: <30s total time spent in $AF_l$[14], Mild AF ($AF_{mild}$): >30 s total time spent in AF and AFB is <4%[14,19], Moderate AF ($AF_{mod}$): AFB is in the range 4-80%, Severe $AF_l$ ($AF_{sev}$): AFB >80%. The AFB was defined as follows[12]:

$$AFB = \frac{\sum_{n=1}^{N} t_i \times \mathbb{I}_i}{\sum_{n=1}^{N} t_i},$$

with $N$ representing the number of available windows, $t_i$ the length of the $i_{th}$ window (ms) and $\mathbb{I}_i$ the unity operator which was equal to 1 for $AF_l$ and otherwise zero.

For a given recording, $E_{AF}(\%)$ was defined as:

$$E_{AF}(\%) = \frac{\sum_{n=1}^{N} t_i \times (\hat{y}_i - y_i)}{\sum_{n=1}^{N} t_i},$$

where $y_i$ is a binary value representing the window label and $\hat{y}_i$ is the binary label predicted by the model. The binary value 1 was used for $AF_l$ and 0 for Non-AF. The $|E_{AF}(\%)|$ was computed for each $AF_l$ target population (Non-AF, $AF_{mild}$, $AF_{mod}$, $AF_{sev}$). The median and interquartiles, as measures of central tendency and dispersion, were reported for $|E_{AF}(\%)|$.

For the simulated intended use scenario, the Se, Sp, PPV and negative predictive value (NPV) performance statistics were calculated on a per patient basis.

**Data availability**

The data that support the findings of this study included raw ECG and manual annotations for atrial tachyarrhythmia as well as demographic information. Data may be made available for noncommercial academic use from the authors with permission from the respective hospitals they originated from. Please contact the corresponding author for such requests.

**Code availability**

Source code for computing the heart rate variability and morphological features used in the XGB model are available on our open source platform at physiozoo.com

**Acknowledgements**

This research was supported for S.B., S.G., and J.A.B. by a grant (3-17550) from the Ministry of Science & Technology, Israel & Ministry of Europe and Foreign Affairs (MEAE) and the Ministry of Higher Education, Research and Innovation (MESRI) of France and ERANET - 3-16881 from the Israeli Ministry of Health. We thank Iris Eisen from the Department of Cardiology at Rambam Health Care Campus and Aya Gosa from the Department of Cardiovascular Medicine, Faculty of Medicine, Saitama Medical University International Medical Center, Japan for supporting the export of the Holter data.

**Author contributions**

SB performed the data analysis (feature engineering, statistical analysis and machine learning models' developments), contributed to the methodology, developed the remote annotation platform and to the writing of the original draft. TS contributed to the ArNet2 algorithm's development. MA, JO and JMS defined the annotation protocol and annotated the test set recordings. EM, MS, KT and AE contributed to the development of the Israeli and Japanese Holter datasets and to the error analysis by reviewing the false positive and false negative recordings. MS supervised the work of EM and AE at Rambam. JB contributed to the conceptualization, methodology, supervision of SB and wrote the original draft. All authors reviewed the manuscript and provided extensive comments.

**Competing interests**

The authors declare no competing interest.

**Correspondence**

Correspondence and requests for materials should be addressed to Joachim A. Behar (jbehar@technion.ac.il)

**References**

1   Björck S, Palaszewski B, Friberg L, Bergfeldt L. Atrial fibrillation, stroke risk, and warfarin therapy revisited: A population-based study. *Stroke* 2013; **44**: 3103–8.

2   Haim M, Hoshen M, Reges O, Rabi Y, Balicer R, Leibowitz M. Prospective national study of the prevalence, incidence, management and outcome of a large contemporary cohort of patients with incident non-valvular atrial fibrillation. *J Am Hear Assoc* 2015; **4**: 1–12.

3   Wolf PA, Abbott RD, Kannel WB. Atrial fibrillation as an independent risk factor for stroke: The framingham study. *Stroke* 1991; **22**: 983–8.

4   Shen AY-J, Contreras R, Sobnosky S, *et al.* Racial/ethnic differences in the prevalence of atrial fibrillation among older adults—a cross-sectional study. *J Natl Med Assoc* 2010; **102**: 906–14.

5   Chugh SS, Havmoeller R, Narayanan K, *et al.* Worldwide epidemiology of atrial fibrillation: a Global Burden of Disease 2010 Study. *Circulation* 2014; **129**: 837–47.

6   Pothineni N V, Vallurupalli S. Gender and atrial fibrillation: differences and disparities. *US Cardiol Rev* 2018; **12**: 24–33.

7   Lindow T, Kron J, Thulesius H, Ljungström E, Pahlm O, Ljungstr€ E. Erroneous computer-based interpretations of atrial fibrillation and atrial flutter in a Swedish primary health care setting. *Taylor Fr* 2019; **37**: 426–33.

8   Carrara M, Carozzi L, Moss TJ, *et al.* Heart rate dynamics distinguish among atrial fibrillation, normal sinus rhythm and sinus rhythm with frequent ectopy. *Physiol Meas* 2015; **36**: 1873–88.

9   Moss TJ, Lake DE, Moorman JR. Local dynamics of heart rate: detection and prognostic implications. *Physiol Meas* 2014; **35**: 1929–42.

10  Goldberger AL, Amaral LAN, Glass L, *et al.* PhysioBank, PhysioToolkit, and PhysioNet: components of a new research resource for complex physiologic signals. *Circulation* 2000; **101**: e215--e220.

11  Wang X, Ma C, Zhang X, Gao H, Clifford G, Liu C. Paroxysmal atrial fibrillation events detection from dynamic ECG recordings: The 4th China physiological signal challenge 2021. In: Proc. PhysioNet. 2021: 1–83.

12  Chocron A, Oster J, Biton S, *et al.* Remote atrial fibrillation burden estimation using deep recurrent neural network. *Under Revis IEEE T Bio-Med* 2020; published online Aug.

13   Wang J. An intelligent computer-aided approach for atrial fibrillation and atrial flutter signals classification using modified bidirectional LSTM network. *Inf Sci (Ny)* 2021; **574**: 320–32.

14   Kirchhof P, Benussi S, Kotecha D, *et al.* 2016 ESC Guidelines for the management of atrial fibrillation developed in collaboration with EACTS. *Kardiol Pol* 2016; **37**: 2893–962.

15   Behar JA, Rosenberg A, Alexandrovich A, *et al.* PhysioZoo: A Novel Open Access Platform for Heart Rate Variability Analysis of Mammalian Electrocardiographic Data. *Front Physiol* 2018; **9**: 1390.

16   Gendelman S, Biton S, Derman R, *et al.* PhysioZoo ECG: Digital electrocardiography biomarkers to assess cardiac conduction. In: 2021 Computing in Cardiology (CinC). 2021: 1–4.

17   Head T, MechCoder GL, Shcherbatyi I, others. scikit-optimize/scikit-optimize: v0. 5.2. *Zenodo* 2018.

18   Go AS, Reynolds K, Yang J, *et al.* Association of burden of atrial fibrillation with risk of ischemic stroke in adults with paroxysmal atrial fibrillation: the KP-RHYTHM study. *JAMA Cardiol* 2018; **3**: 601–8.

19   Boriani G, Glotzer T V., Ziegler PD, *et al.* Detection of new atrial fibrillation in patients with cardiac implanted electronic devices and factors associated with transition to higher device-detected atrial fibrillation burden. *Hear Rhythm* 2018; **15**: 376–83.

20   Lake DE, Moorman JR. Accurate estimation of entropy in very short physiological time series: the problem of atrial fibrillation detection in implanted ventricular devices. *Am J Physiol Circ Physiol* 2011; **300**: H319--H325.

21   Sarkar S, Ritscher D, Mehra R. A detector for a chronic implantable atrial tachyarrhythmia monitor. *IEEE Trans Biomed Eng* 2008; **55**: 1219–24.

22   Granada J, Uribe W, Chyou P-H, *et al.* Incidence and predictors of atrial flutter in the general population. *J Am Coll Cardiol* 2000; **36**: 2242–6.

23   Svendsen JH, Diederichsen SZ, Højberg S, *et al.* Implantable loop recorder detection of atrial fibrillation to prevent stroke (The LOOP Study): a randomised controlled trial. *Lancet* 2021; **398**: 1507–16.

24   Mohanty S, Mohanty P, Di Biase L, *et al.* Results from a single-blind, randomized study comparing the impact of different ablation approaches on long-term procedure outcome in coexistent atrial fibrillation and flutter (APPROVAL). *Circulation* 2013; **127**: 1853–60.

25   Celikyurt U, Knecht S, Kuehne M, *et al.* Incidence of new-onset atrial fibrillation after cavotricuspid isthmus ablation for atrial flutter. *Ep Eur* 2017; **19**: 1776–80.

26   Link MS. Evaluation and initial treatment of supraventricular tachycardia. *N Engl J Med* 2012; **367**: 1438–48.

27   Link MS. Introduction to the arrhythmias: a primer. *EP Lab Dig* 2007; **5**: 38–9.

28   He K, Zhang X, Ren S, Sun J. Deep residual learning for image recognition. In: Proceedings of the IEEE conference on computer vision and pattern recognition. 2016: 770–8.

29  Kingma DP, Ba J. Adam: A method for stochastic optimization. *arXiv Prepr arXiv14126980* 2014.

**List of Tables**

|  | **UVAF-test** | **SHDB-test** | **CPSC-test** | **RBDB-test** | **RBDB-test2** |
|---|---|---|---|---|---|
| Origin | USA | Japan | China | Israel | Israel |
| Patients, n | 100 | 100 | 102 | 100 | 1,608 |
| Age (yrs) $\mu$ (Q1 − Q3) | 69.0 (58.8-76.2) | 69.5 (62.0-75.2) | 69.0 (59.0-76.0) | 69.5 (56.8-78.0) | 67.0 (50.0-77.0) |
| Female, n | 50 (50%) | 45 (45%) | 41 (40%) | 50 (50%) | 811 (43.7%) |
| Recordings length (sec) $\mu$ (Q1 − Q3) | 24.0, (24.0-24.0) | 24.00, (23.98-24.0) | 0.17, (0.08-0.50) | 24.17, (21.70-25.17) | 23.98, (22.02-25.17) |

**Table 1:** Description of the population sample included in the internal test set from UVAF-test and external test sets (RBDB-test, SHDB-test and CPSC-test) and for the intended case scenario test set RBDB-test2. Median ($\mu$) and the interquartile range (Q1-Q3) are reported for age and recording length. The University of Virginia dataset (UVAF) from the USA, the Saitama Hospital dataset (SHDB) from Japan, the China Physiological Signal Challenge (CPSC) from China and Rambam Hospital dataset (RBDB) from Israel.

|  |  | $F_1$ | AUROC | Se | Sp | PPV | $\|E_{AF}(\%)\|$ $\mu$ (Q1 − Q3) |
|---|---|---|---|---|---|---|---|
| Ethnicities | UVAF-test | 0.95 | 0.99 | 093 | 0.96 | 0.96 | 0.10 (0.0-2.41) |
| | SHDB-test | 0.92 | 0.99 | 0.89 | 0.98 | 0.94 | 0.59 (0.10-3.23) |
| | RBDB-test | 0.90 | 0.99 | 0.83 | 0.98 | 0.97 | 0.21 (0.09-7.59) |
| | CPSC-test | 0.95 | 0.99 | 0.95 | 0.97 | 0.94 | 0.0 (0.0-6.02) |
| | Combined test set | 0.92 | 0.99 | 0.89 | 0.97 | 0.96 | 0.32 (0.01-3.62) |
| Genders | Female | **0.94** | **0.99** | **0.92** | **0.98** | **0.97** | **0.14 (0.0-2.46)** |
| | Male | 0.90 | 0.98 | 0.86 | 0.97 | 0.95 | 0.71 (0.06-4.79) |
| Ages | ≤ 60 | **0.95** | **1.00** | **0.93** | **0.99** | **0.97** | **0.14 (0.0-2.81)** |
| | 60 to 75 | 0.90 | 0.98 | 0.86 | 0.97 | 0.96 | 0.47 (0.05-5.15) |
| | ≥ 75 | 0.94 | 0.98 | 0.91 | 0.94 | 0.96 | 0.22 (0.0-2.66) |

**Table 2:** Performance measures of ArNet2 across ethnicities, genders and ages.

**List of figures**

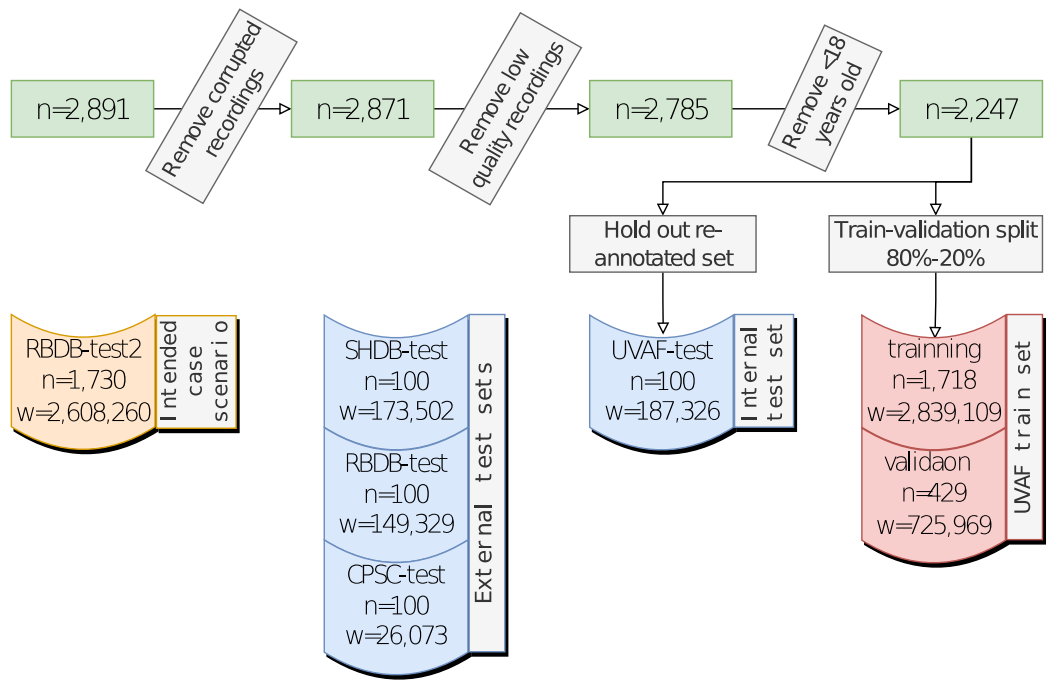

**Figure 1:** Data exclusion and stratification process applied to the UVAF database and description of all datasets used in this research.

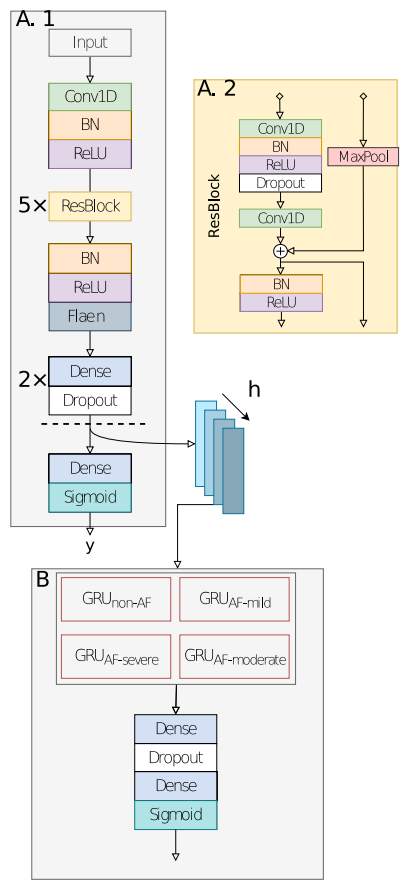

**Figure 2:** ArNet2 architecture. A deep CNN with ResNet blocks (ResBlock) is trained at the first stage, (a.1). The ResBlock architecture given in a.2. This first stage network is trained for the task of binary classification on 60-beat random windows. Before the last layer, the embedded features are extracted. The second stage consists of an RNN with GRU units (b). Its inputs are the extracted embedded features from the previously trained CNN, both from the current window and its $h$ preceding windows (when available), all being temporally concatenated. Depending on the $AF_l$ severity label predicted by the CNN, a GRU layer is chosen from the pool, which takes as input the window to predict as well as $h$ previous ones. It then goes through two dense layers and outputs a probability label of $AF_l$ for each window.

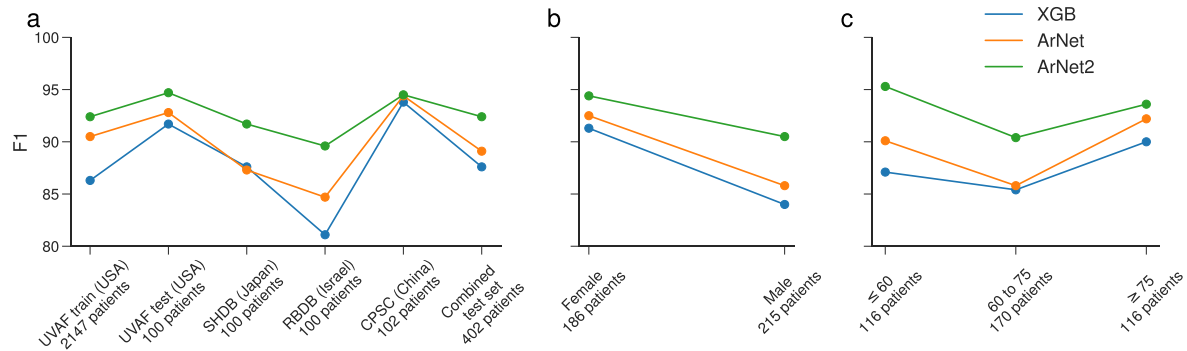

**Figure 3:** F1 performance for the models (XGB, ArNet and ArNet2) in classifying $AF_l$ events across ethnicity (panel a), sex (panel b) and age (panel c).

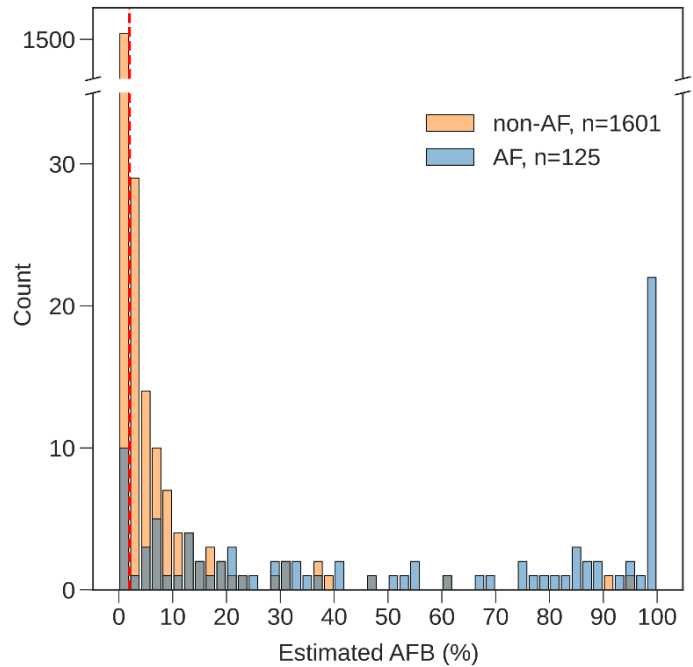

**Figure 4**: Histogram of the ArNet2 estimated AFB for the RBAF-test2 dataset (n=1,726) simulating the intended use scenario.

**Supplementary figures**

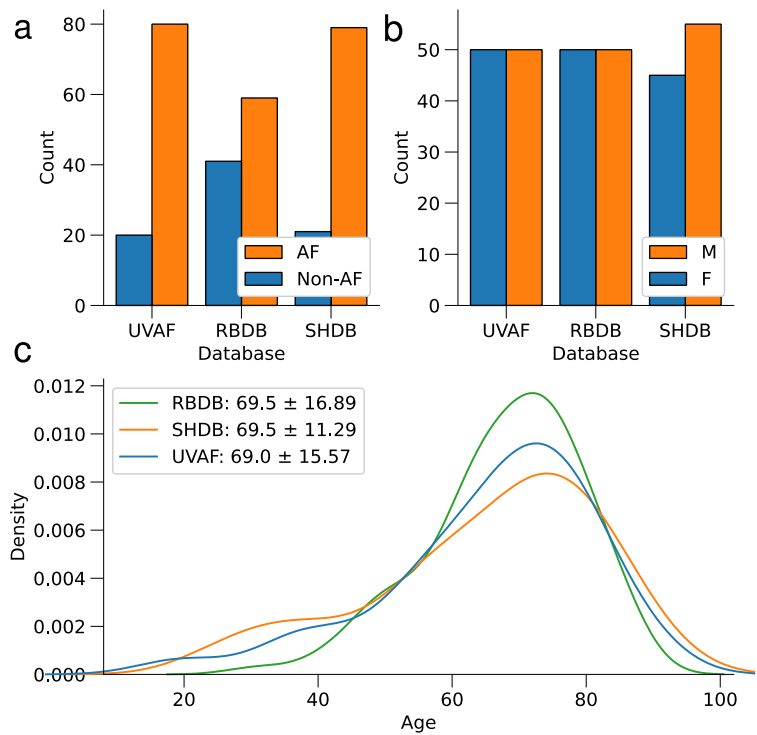

**Figure S1:** Distribution for the re-annotated test sets. Label distribution (panel a), sex distribution (panel b) and age distribution (panel c) over the three test sets.

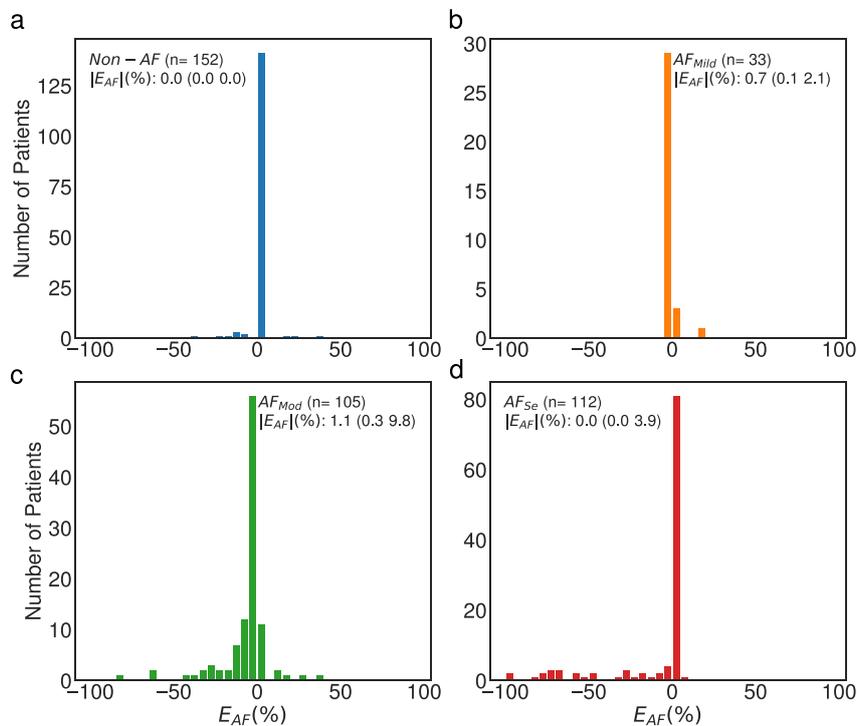

**Figure S2:** $|E_{AF}(\%)|$ for the combined test set (n=402) per different $\mathbf{AF}_l$ severity labels; Non-AF (panel a), $\mathbf{AF}_{mild}$ (Panel b), $\mathbf{AF}_{mod}$ (panel c) and $\mathbf{AF}_{sev}$ (panel d).

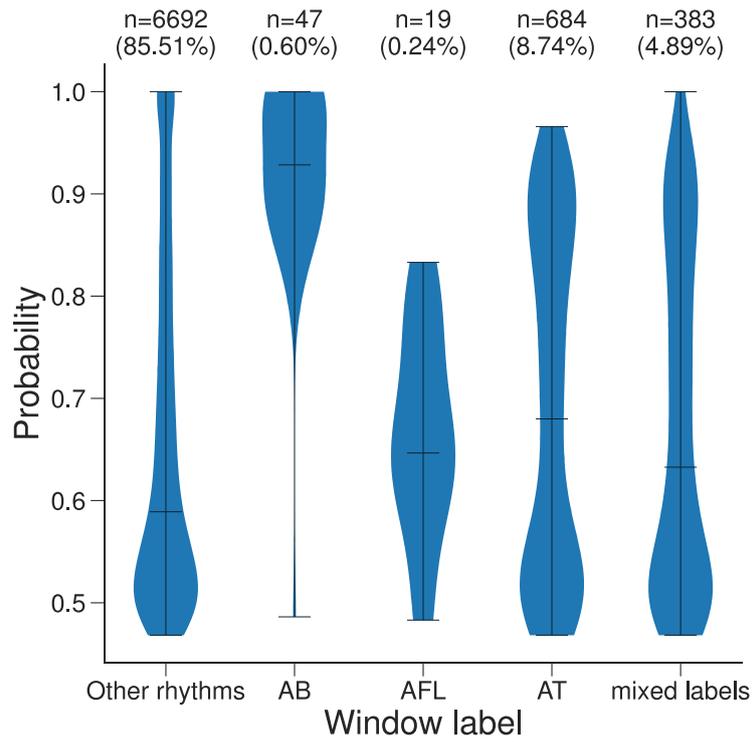

**Figure S3:** The probability output of ArNet2 for the combined test set FP grouped to the true window labels. The represented rhythms shown in the violin plots are Other rhythms, Atrial Bigeminy (AB), Atrial Tachycardia (AT) and mixed labels. The mixed labels group consists of windows with $AF_l$ and other dominant rhythm so that the label of this window is non-AF.

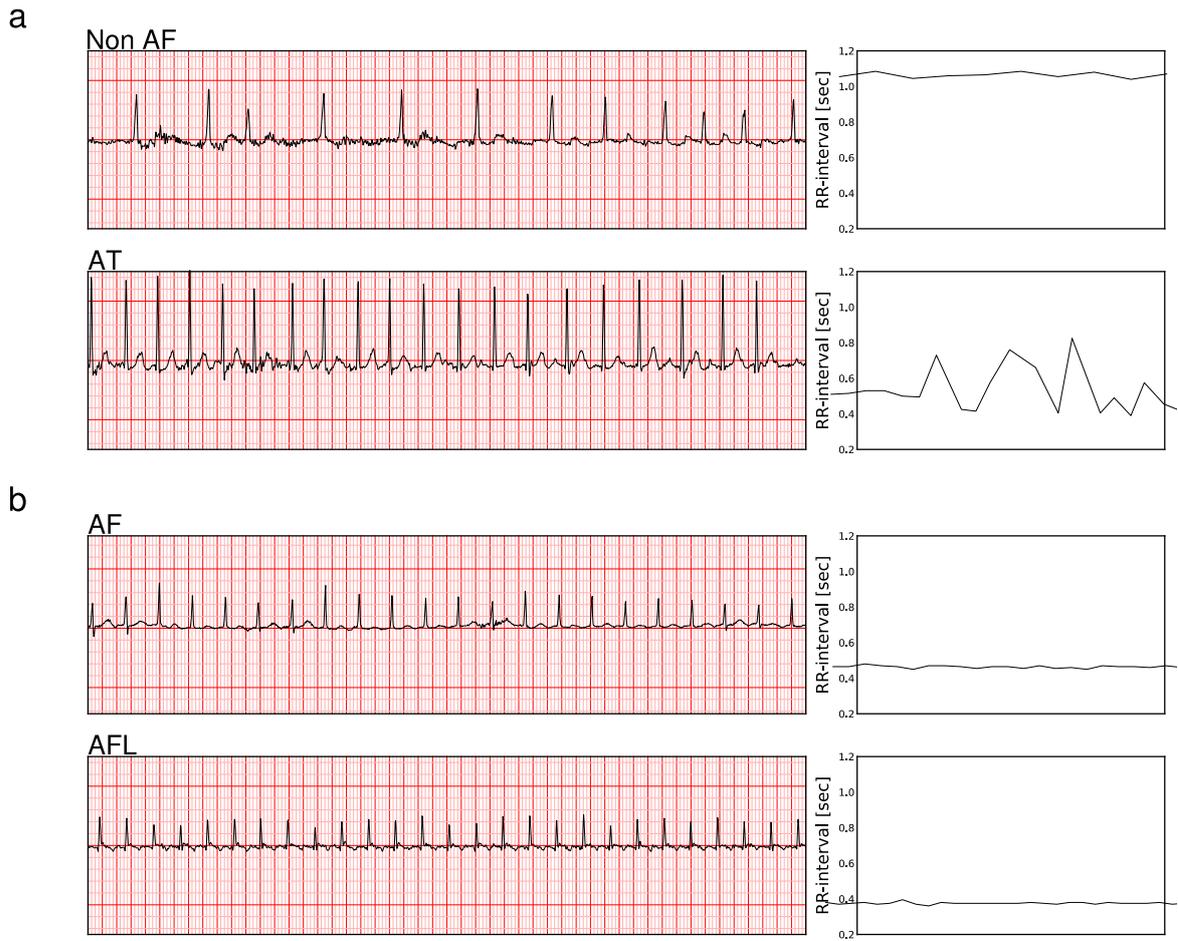

**Figure S4:** Examples of misclassified windows where a representative short 10-sec of each window is presented followed by the window RR time series. FP examples (panel a) and FN (panel b).

**Supplementary Tables**

| Hyperparameter | Type | Range | Prior |
|---|---|---|---|
| $w_s$ | Categorical | [60, 70, 80, 90, 100, 110, 120] | N.A. |
| $n_b$ | Integer | [3, 7] | Uniform |
| $n_f$ | Integer | $[2^5, 2^7]$ | Log uniform |
| $f_l$ | Integer | [3, 10] | Uniform |
| $d_{r1}$ | Real/Continuous | [0, 0.5] | Uniform |
| $n_{h_u}$ | Integer | $[2^6, 2^9]$ | Log uniform |
| $d_{r2}$ | Real/Continuous | [0, 0.8] | Uniform |
| $\alpha$ | Real/Continuous | $[10^{-5}, 10^{-2}]$ | Log uniform |

**Table S1:** Hyperparameter search space for Bayesian optimization. for Uniform and Log uniform distribution the range is given with (lower, upper) bounds.

| | | $F_1$ | AUROC | Se | Sp | PPV | $\|E_{AF}(\%)\|$ $\mu$ (Q1 − Q3) |
|---|---|---|---|---|---|---|---|
| UVAF-train | XGB | 0.86 | 0.99 | 0.85 | 0.99 | 0.88 | 2.43 (0.52-11.29) |
| | ArNet | 0.90 | 0.97 | 0.87 | **1.00** | 0.94 | **0.10 (0.0-6.27)** |
| | ArNet2 | **0.92** | **1.00** | **0.93** | 0.99 | 0.91 | 0.15 (0.0-4.02) |
| UVAF-test | XGB | 0.92 | 0.97 | 0.88 | 0.96 | 0.96 | 1.42 (0.30-5.31) |
| | ArNet | 0.93 | 0.98 | 0.89 | 0.97 | **0.97** | **0.08 (0.0-3.29)** |
| | ArNet2 | **0.95** | **0.99** | **0.93** | 0.96 | 0.96 | 0.10 (0.0-2.41) |
| SHDB-test | XGB | 0.88 | 0.97 | 0.82 | 0.98 | 0.94 | **0.44 (0.13-4.86)** |
| | ArNet | 0.87 | 0.97 | 0.81 | **0.99** | **0.95** | 0.81 (0.23-3.73) |
| | ArNet2 | **0.92** | **0.99** | **0.89** | 0.98 | 0.94 | 0.59 (0.10-3.23) |
| RBDB-test | XGB | 0.81 | 0.95 | 0.70 | 0.98 | 0.97 | 1.59 (0.27-32.61) |
| | ArNet | 0.85 | 0.98 | 0.74 | **0.99** | **0.99** | 0.36 (0.11-14.94) |
| | ArNet2 | **0.90** | **0.99** | **0.83** | 0.98 | 0.97 | **0.21 (0.09-7.59)** |
| CPSC-test | XGB | 0.94 | 0.97 | 0.92 | 0.98 | 0.96 | 5.06 (0.62-19.03) |
| | ArNet | **0.95** | 0.97 | 0.91 | **0.99** | **0.99** | 0.0 (0.0-12.78) |
| | ArNet2 | **0.95** | **0.99** | **0.95** | 0.97 | 0.94 | **0.0 (0.0-6.02)** |
| Combined test set | XGB | 0.88 | 0.97 | 0.81 | **0.98** | 0.96 | 1.33 (0.25-9.09) |
| | ArNet | 0.89 | 0.98 | 0.82 | **0.98** | **0.97** | 0.39 (0.02-6.11) |
| | ArNet2 | **0.92** | **0.99** | **0.89** | 0.97 | 0.96 | **0.32 (0.01-3.62)** |

**Table S2:** Performance statistics for the UVAF-train, UVAF-test and external test sets, and the overall combined test set.

|  |  | F$_1$ | AUROC | Se | Sp | PPV | $\|E_{AF}(\%)\|$ μ (Q1 − Q3) |
|---|---|---|---|---|---|---|---|
| Female | XGB | 0.91 | 0.98 | 0.86 | 0.99 | 0.98 | 1.14 (0.27-9.10) |
|  | ArNet | 0.92 | 0.99 | 0.87 | **0.99** | **0.98** | 0.26 (0.0-2.67) |
|  | ArNet2 | **0.94** | **0.99** | **0.92** | 0.98 | 0.97 | **0.14 (0.0-2.46)** |
| Male | XGB | 0.84 | 0.95 | 0.76 | 0.97 | 0.94 | 1.46 (0.21-9.10) |
|  | ArNet | 0.86 | 0.86 | 0.97 | 0.78 | **0.98** | 0.80 (0.07-6.96) |
|  | ArNet2 | **0.90** | **0.98** | **0.86** | 0.97 | 0.95 | **0.71 (0.06-4.79)** |

**Table S3:** Performance statistics for the combined test set divided by sex.

|  |  | F$_1$ | AUROC | Se | Sp | PPV | $\|E_{AF}(\%)\|$ μ (Q1 − Q3) |
|---|---|---|---|---|---|---|---|
| ≤ 60 | XGB | 0.87 | 0.98 | 0.80 | 0.99 | 0.94 | 0.67 (0.19-4.27) |
|  | ArNet | 0.90 | 0.99 | 0.83 | **1.00** | **0.99** | 0.26 (0.0-3.62) |
|  | ArNet2 | **0.95** | **1.00** | **0.93** | 0.99 | 0.97 | **0.14 (0.0-2.81)** |
| 60 to 75 | XGB | 0.86 | 0.96 | 0.77 | **0.98** | **0.97** | 1.13 (0.16-10.63) |
|  | ArNet | 0.86 | 0.97 | 0.77 | **0.98** | **0.97** | 0.89 (0.09-9.89) |
|  | ArNet2 | **0.90** | **0.98** | **0.86** | 0.97 | 0.96 | **0.47 (0.05-5.15)** |
| ≥ 75 | XGB | 0.90 | 0.95 | 0.85 | 0.94 | 0.96 | 2.03 (0.61-9.09) |
|  | ArNet | 0.92 | **0.98** | 0.88 | **0.96** | **0.97** | 0.24 (0.0-3.81) |
|  | ArNet2 | **0.94** | 0.98 | **0.91** | 0.94 | 0.96 | **0.22 (0.0-2.66)** |

**Table S4**: Performance statistics for the combined test set grouped into three different age groups.

|  |  | **Min** | **Q1** | **Median** | **Q3** | **Max** |
|---|---|---|---|---|---|---|
| UVAF-train | XGB | 0.0 | 0.52 | 2.43 | 11.29 | 95.68 |
|  | ArNet | 0.0 | 0.0 | 0.1 | 6.27 | 100.0 |
|  | ArNet2 | 0.0 | 0.0 | 0.15 | 4.02 | 80.56 |
| UVAF-test | XGB | 0.0 | 0.30 | 1.42 | 5.31 | 81.73 |
|  | ArNet | 0.0 | 0.0 | 0.08 | 3.29 | 83.56 |
|  | ArNet2 | 0.0 | 0.0 | 0.10 | 2.41 | 90.04 |
| SHDB-test | XGB | 0.0 | 0.13 | 0.44 | 4.86 | 72.67 |
|  | ArNet | 0.03 | 0.23 | 0.81 | 3.73 | 70.78 |
|  | ArNet2 | 0.0 | 0.10 | 0.59 | 3.23 | 43.54 |
| RBDB-test | XGB | 0.0 | 0.27 | 1.59 | 32.61 | 97.00 |
|  | ArNet | 0.02 | 0.11 | 0.36 | 14.94 | 97.79 |
|  | ArNet2 | 0.0 | 0.09 | 0.21 | 7.59 | 99.20 |

|  |  | Min | Q1 | Median | Q3 | Max |
|---|---|---|---|---|---|---|
| CPSC-test | XGB | 0.0 | 0.62 | 5.06 | 19.03 | 53.76 |
| | ArNet | 0.0 | 0.0 | 0.0 | 12.78 | 67.67 |
| | ArNet2 | 0.0 | 0.0 | 0.0 | 6.02 | 48.71 |
| Combined test set | XGB | 0.0 | 0.25 | 1.33 | 9.09 | 97.00 |
| | ArNet | 0.0 | 0.02 | 0.39 | 6.11 | 97.79 |
| | ArNet2 | 0.0 | 0.01 | 0.32 | 3.62 | 99.20 |

**Table S5:** $|E_{AF}(\%)|$ statistics across all ethnicity groups, i.e for the UVAF train and test sets, and the generalization performance over all external test sets.

|  |  | Min | Q1 | Median | Q3 | Max |
|---|---|---|---|---|---|---|
| Female | XGB | 0.0 | 0.27 | 1.14 | 9.10 | 80.99 |
| | ArNet | 0.0 | 0.00 | 0.26 | 2.67 | 96.18 |
| | ArNet2 | 0.0 | 0.00 | 0.14 | 2.46 | 99.20 |
| Male | XGB | 0.0 | 0.21 | 1.46 | 9.10 | 97.0 |
| | ArNet | 0.0 | 0.07 | 0.80 | 6.96 | 97.79 |
| | ArNet2 | 0.0 | 0.06 | 0.71 | 4.79 | 93.36 |

**Table S6:** $|E_{AF}(\%)|$ statistics for combined test set stratified by sex.

|  |  | Min | Q1 | Median | Q3 | Max |
|---|---|---|---|---|---|---|
| ≤ 60 | XGB | 0.0 | 0.19 | 0.67 | 4.27 | 92.52 |
| | ArNet | 0.0 | 0.0 | 0.26 | 3.62 | 84.31 |
| | ArNet2 | 0.0 | 0.0 | 0.14 | 2.81 | 41.92 |
| 60 to 75 | XGB | 0.0 | 0.16 | 1.13 | 10.63 | 82.84 |
| | ArNet | 0.0 | 0.09 | 0.89 | 9.89 | 96.18 |
| | ArNet2 | 0.0 | 0.05 | 0.47 | 5.15 | 99.20 |
| ≥ 75 | XGB | 0.0 | 0.61 | 2.03 | 9.09 | 97.0 |
| | ArNet | 0.0 | 0.0 | 0.24 | 3.81 | 97.79 |
| | ArNet2 | 0.0 | 0.0 | 0.22 | 2.66 | 93.36 |

**Table S7:** $|E_{AF}(\%)|$ statistics for combined test set grouped into 3 different age groups.